\documentclass[preprint,authoryear,5p]{elsarticle}
\usepackage[export]{adjustbox}
\usepackage{lineno,hyperref}
\usepackage{dirtytalk}
\usepackage{tikz}
\usepackage{tikz-qtree}
\usepackage{color}
\usepackage{mathtools}
\usepackage{comment}
\usepackage{rotating}
\usepackage{cancel}
\usepackage{algorithm}
\usepackage{algpseudocode}
\usepackage{multirow}
\usepackage{multicol}
\usepackage{graphicx}
\usepackage{booktabs}
\usepackage{listings}
\modulolinenumbers[5]

\journal{Expert Systems With applications}
\biboptions{authoryear}
\begin{document}

\begin{frontmatter}

\title{A Library for Automatic Natural Language Generation of Spanish Texts}

\author[mymainaddress]{Silvia Garc\'ia-M\'endez\corref{mycorrespondingauthor}}
\ead{sgarcia@gti.uvigo.es}
\author[mymainaddress]{Milagros Fern\'andez-Gavilanes}
\ead{mfgavilanes@gti.uvigo.es}
\author[mymainaddress]{Enrique Costa-Montenegro}
\ead{kike@gti.uvigo.es}
\author[mymainaddress]{Jonathan Juncal-Mart\'inez}
\ead{jonijm@gti.uvigo.es}
\author[mymainaddress]{F. Javier Gonz\'alez-Casta\~no}
\ead{javier@det.uvigo.es}
\address[mymainaddress]{GTI Research Group, Telematics Engineering Department, University of Vigo, EI Telecomunicaci\'on, Campus, 36310 Vigo, Spain}
\cortext[mycorrespondingauthor]{Corresponding author: sgarcia@gti.uvigo.es}

\begin{abstract}
In this article we present a novel system for {\em natural language generation} ({\sc nlg}) of Spanish sentences from a minimum set of meaningful words (such as nouns, verbs and adjectives) which, unlike other state-of-the-art solutions, performs the {\sc nlg} task in a fully automatic way, exploiting both knowledge-based and statistical approaches. Relying on its linguistic knowledge of vocabulary and grammar, the system is able to generate complete, coherent and correctly spelled sentences from the main word sets presented by the user. The system, which was designed to be integrable, portable and efficient, can be easily adapted to other languages by design and can feasibly be integrated in a wide range of digital devices. During its development we also created a supplementary lexicon for Spanish, {\em aLexiS}, with wide coverage and high precision, as well as syntactic trees from a freely available definite-clause grammar. The resulting {\sc nlg} library has been evaluated both automatically and manually (annotation). 
The system can potentially be used in different application domains such as augmentative communication and automatic generation of administrative reports or news.
\end{abstract}


\end{frontmatter}

\section{Introduction}
\label{intro}
Natural language generation ({\sc nlg}) has attracted increasing interest in the field of human-computer interaction, as it responds to the demand for coherent and natural-sounding fully machine-generated texts. {\sc nlg} was for some time considered a sub-field of natural language processing ({\sc nlp}). However, due to its growing significance and the fact that it requires expertise in various research areas, including linguistics and computation, it has evolved into a major research topic and a discipline in its own right. It has been defined as \say{{\em [...]the sub-field of artificial intelligence ({\sc ai}) and computational linguistics that focuses on computer systems that can produce understandable texts in English and other human languages, typically starting from some non-linguistic representation of information as input[...]}}~\citep{ReiterDale00}. Nevertheless, this definition may be considered obsolete since, as we will explain later, the input to {\sc nlg} systems consists not only of non-linguistic information like objective data, but also linguistic information (words, sentences, texts) and even visual data.

Traditionally, {\sc nlg} focused on text-to-text generation, regarding which many sub-fields existed, such as summarizing, text simplification and automatic question generation. The earliest systems took inputs like words, sentences and even whole texts to produce new text as output. However, new data-driven methods have expanded the possibilities of {\sc nlg}. Most data-to-text generation methods rely on predefined templates to automatically transform data into text by filling gaps in predefined text templates, which has applications in reportage of weather, traffic, sports, health, etc. The more recent vision-to-text systems~\citep{thomason14} produce texts, mainly using deep-learning approaches, from visual representations of information such as pictures. 
 
It is broadly agreed that {\sc nlg} has only recently begun to take full advantage of recent advances in data-driven, machine learning and deep-learning techniques.

{\sc nlg} tasks are generally addressed by splitting them into sub-problems~\citep{ReiterDale00, ReiterDale1997}: content determination (deciding which events are important), text structuring (ordering information in the output text), sentence aggregation, lexicalization (finding the right words and sentences to express information), referring expression generation (identification of domain objects) and linguistic realization (generation of well-formed texts). New general systems and applications follow these trends in a broad sense, described in more detail in what follows.

Content determination consists of selecting which information should be included and excluded in the process of text generation. It may be seen as a filtering process and is clearly context/application-dependent. The result is abstracted into semantic, formal, logical and graph structures. Researchers have started exploring data-driven techniques for this sub-problem~\citep{kutlak13}. Text structuring and discourse planning is the ordering of sentences or paragraphs in how they are presented to readers. The importance of individual events (sentence/paragraph) for the final audience is assessed, taking into consideration internal relations, strongly dependent on the application domain. State-of-the-art solutions include manual rule-based approaches \citep{mairesse2007personage,duvsek2015training} and rhetorical structure theory ({\sc rst}) \citep{williams2008generating}, machine learning techniques \citep{lampouras2016imitation,mei2015talk}, systemic functional grammar ({\sc sfg}) \citep{Bateman97}, meaning-text theory ({\sc mtt}) \citep{wanner2010marquis} and the centred theory/approach \citep{barzilay2008modeling}, among others. Sentence aggregation at the semantic or syntactical level, which tries to join data into single sentences, deals with fluency and readability; however, conceptually the approach is complex. Lexicalization converts the result of the previous stage into natural language ({\sc nl}) but has the problem that there may exist many ways to express the same idea in {\sc nl}. Logically, the more possibilities the system can choose from, the better. While a simple approach is to convert domain concepts into lexical items, sometimes the task is made more complex due to gradable properties, such as size and colour. Referring expression generation ({\sc reg}) consists of selecting the words or phrases that can describe domain entities in an unambiguous manner. Selected is the best set of known properties to distinguish an element from others, while any information that is not directly relevant to the identification task is discarded. Several algorithms can be found in the literature for this purpose including the full brevity procedure \citep{dale1989cooking}, the greedy heuristic algorithm \citep{dale1992generating,frank2009using} and the incremental algorithm \citep{dale1995computational}. Finally, linguistic realization involves text ordering, the generation of morphological forms, the insertion of function words like auxiliary verbs and prepositions and the insertion of punctuation marks. Here the {\em generation gap} appears, because sometimes it is necessary to add elements that are not present in the input data. Templates are widely used to address this issue but only when the domain is small and the expected variation is minimal. These approaches yield better quality than other approaches but are very time consuming and do not scale well enough in certain applications with high linguistic variation. Hand-coded grammar-based systems are the alternative to templates but they require very detailed input such as {\sc kpml}~\citep{Bateman97} based on systemic-functional grammar ({\sc sfg}). Other alternatives include statistical approaches, which derive probabilistic grammar from large corpora, reducing effort while increasing coverage~\citep{Langkilde02}; the head-driven phrase structure grammar ({\sc hpsg}) \citep{nakanishi2005probabilistic}, the lexical-functional grammar ({\sc lfg}) \citep{cahill2006robust} and the tree-adjoining grammar ({\sc tag}) \citep{gardent2015multiple}. We follow a hybrid approach that exploits the advantages of grammar-based and stochastic systems but also reduces the effort of the {\sc nlg} process.

As noted in~\citet{Stent05}, generated quality depends on adequacy, fluency, readability and variation. We can distinguish three main {\sc nlg} architectures: traditional modular architectures (macro-planning or selection and text structuring, micro-planning or sentence aggregation, lexicalization, referring expression generation, and linguistic realization applying syntactic and morphological rules), planning perspective architectures (less modular but also with roots in the {\sc ai} tradition) and data-driven approaches. The latter rely on statistical learning and represent a strong trend in {\sc nlg}, but the most widely adopted approach today is the rule-based (or template-based) method ~\citep{cheyer2007method,mirkovic08}.

The trade-off between output quality and efficiency is becoming a central issue. Recent years have witnessed a marked interest in automatic text generation, where ``automatic" means that the user is only required to introduce meaningful words like nouns, verbs and adjectives.

We are interested in automatically generating complete, coherent and correctly spelled sentences from specific content specified by the user at word level as ``key points" (verbs, nouns and adjectives). Our practical approach to text-to-text generation is easily adaptable to other languages and integrable in a wide range of digital devices. This work is not our first attempt at developing an automatic {\sc nlg} system for Spanish. In~\citet{silvia2018web4all} we described an automatic version of SimpleNLG for Spanish but soon realized that it was difficult to expand and improve. This new automatic {\sc nlg} system is based on a modular architecture that allows domain-dependent components to be separated from domain independent components.

The rest of this paper is organized as follows. First we review the state-of-the-art for {\sc nlg} (Section ~\ref{related_work}). We then describe a hybrid system combining linguistic rule-based and statistical techniques, composed of two subsystems: a knowledge base consisting of the {\em aLexiS} Spanish lexicon (Section~\ref{lexicon}) and a grammar (Section~\ref{grammar}). We then describe an interface engine (Section~\ref{nlg_system}), followed by our evaluation results (Section ~\ref{evaluacion}). To test the library we created a dataset specially tailored for automatic {\sc nlg} (Section~\ref{corpus}) and used manual and automatic procedures (Section~\ref{evaluation}). We compared our system with an automatic Spanish version of the SimpleNLG library~(Section \ref{simpleNLG}). Finally, in Section~\ref{conclusions} we conclude the paper.

\section{Related work}
\label{related_work}

{\sc nlg} started in the second half of the 20th century with automatic translation~\citep{Sager67}. Research in the 1970s focused on choosing appropriate words to express abstract conceptual content and using these to generate appropriate textual structures~\citep{Schank75,Mann82}. It was not until the 1980s when {\sc nlg} was truly recognized beyond language understanding in reverse mode. A number of significant developments occurred during this decade~\citep{McKeown85,Appelt85}, marked by the move away from large monolithic systems that attempted to resolve all {\sc nlg} stages of specific problems. However, by the end of the 1980s substantial research adopted an {\sc ai} perspective. During the 1990s there were significant new achievements~\citep{ReiterDale1997,HOVY93} and the first real-world {\sc nlg} applications appeared, including the pioneering FoG weather forecasting system~\citep{Goldberg94}. In addition, the interest in multilingual generation grew. 

Nowadays, {\sc nlg} systems are considered to be highly sensitive to problem characterization and, in most cases, are purpose-built. Conversely, our system is based on a modular architecture that allows us to separate domain-dependent (grammar and lexica) from domain-independent components ({\sc nlg} surface realizer). This means that it can be adapted to different purposes and fields of interest using specific syntactic structures and vocabulary.

From a practical perspective, {\sc nlg} is capable of generating rich and coherent texts, nearly indistinguishable from those created by humans, that is, satisfying completeness (containing enough meaningful information), grammatical and orthographic correctness and semantic coherence criteria. 

If we consider generated text types, we can mainly distinguish between informative texts, summaries, simplified texts, persuasive texts, dialogues, recommendations and creative texts. 

The systems to generate informative texts from objective data, e.g. SumTime~\citep{Reiter05}, which does not perform text-to-text generation and is only available for English, are highly language-sensitive and so are only suitable for the language they were designed for. Most of these systems generate routine documents~\citep{Reiter95}, such as administrative documents, response letters, operating manuals, weather forecasts~\citep{Belz08} and traffic and academic reports. At the other extreme, more creative texts are more challenging to generate. Experiments in this line based on predefined templates are very rigid~\citep{Peinado04}. We seek to develop a tool that should not constrain the content of the generated sentences but assist users to express themselves with the words that come to mind without loss of application generality. Affective {\sc nlg} systems can, in fact, generate texts beyond factual information to create, among other possibilities, poetry~\citep{Gervas2001}, but this goes beyond our purposes and, with these systems, users have no control over content generation.

Summary generation has applications in medicine, finance and sport. Persuasive texts try to influence user behaviours; a representative example is STOP~\citep{Reiter03} to discourage smoking. Dialogue systems focus on human-machine communication and are of interest for call centres or games~\citep{Koller10}. Some aim at generating explanations as sequences of steps (for instance, P.REX~\citep{Fiedler05}, a tool to generate demonstrations of algorithms). Most of these proposals, like the Shed~\citep{Cheng05} recommender for personalized nutritional plans and the text-to-text generator of Wikipedia articles from Internet documents in~\citet{Sauper09}, require an infeasible amount of time and resource-consuming predefined templates to address generality, which also leads to lack of control over content generation. Flights \citep{White10} is an example of a hybrid tool, which uses templates to organize the input data and the OpenCCG\footnote{Available at {\tt http://openccg.sourceforge.net/}.} framework to generate the final user-oriented text, relying on $n$-gram models and language factorized grammars ({\sc flm}). Nevertheless, one of its drawbacks is that it is too specific due to its training process and complex specification of input data. Summing up, templates are still the basis for the main {\sc nlg} approaches, except for some recent systems that apply statistical techniques such as deep learning and that are highly domain-dependent.

We conclude this review by mentioning the most relevant systems nowadays.
The SimpleNLG library (Gatt \& Reiter, 2009) conducts a surface realization task using a knowledge-based approach. It is available for English, with versions also for Brazilian Portuguese (De Oliveira \& Sripada, 2014), French (Vaudry \& Lapalme, 2013), German (Bollmann, 2011) and Italian (Mazzei et al., 2016). This library has influenced the {\sc nlg} field strongly due to its simple input requirements compared to other systems. Its main drawback is that it is not automatic. NaturalOWL (Androutsopoulos
et al., 2014) is a data-to-text tool that generates texts from an ontology ({\sc owl}) (it does not perform text-to-text generation, just data-to-text generation). In this line, in~\citet{dale2005using} the authors presents a tool to generate route descriptions. Unlike SimpleNLG, the input format happens to be very complex, defined by a ``route planning markup language". The method in~\citet{liu2017automatic} to create radiology reports automatically relies on templates. Conversely, the system in~\citet{chen2002towards} is composed of a trainable sentence planner and a stochastic surface realizer, similar to our approach, but only available for English.

To the best of our knowledge no other system performs text-to-text generation in an automatic way, regardless of the target language (Spanish or other). Shed and the system by~\cite{Gervas2001} do not allow the user to control content generation. Others like OpenCCG cannot be easily used for other than the purpose for which they were designed or have a complex interface for data input.

In~\citet{silvia2018web4all} we described our automatic Spanish version of SimpleNLG, but, since it was difficult to improve, we created an automatic {\sc nlg} system based on a modular architecture. Furthermore, without loss of generality, we identified a new use case for {\sc nlg} to help people diagnosed with communication disorders such as autism spectrum disorder ({\sc asd}) to express themselves more easily and quickly. The pictograms in their personal communicators may represent concepts or single words, and the users determine the key words that serve as input to create a coherent and complete sentence. In this use case a template-oriented approach would be unacceptably limiting.

\section{Methodology and system architecture}
\label{methodology}

Below we explain our methodology and the resulting system architecture. Firstly, we focus on the morphological part, i.e. the lexicon that provides the system with the indispensable linguistic knowledge (Section~\ref{lexicon}). We then analyse the syntactic stage, which uses a define-clause grammar ({\sc dcg}) for syntactic structuring (Section~\ref{grammar}), and we conclude with an overview of the system (Section~\ref{nlg_system}). 

\subsection{The {\em aLexiS} lexicon}
\label{lexicon}

Since we want our {\sc nlg} system to be fully automatic, not only in selecting the appropriate grammar structure for the words given by the user but also in the inflection of these words, we need a wide vocabulary and its corresponding linguistic data. This is why we created the {\em aLexiS} lexicon, achieved by interpreting input resources and merging lexica following two well-defined steps~\citep{CrouchKing05} with an intermediate verification step. The merging process is automatic, that is, without human supervision, and proceeds as follows:

\begin{enumerate}
\item Extraction and mapping: all possible entries are extracted and mapped to a common format.
\item Verification: all extracted and mapped entries are verified at a lexical level to check if they are commonly accepted in the {\em Diccionario de la Real Academia de la Lengua Espa\~nola}\footnote{Available at {\tt http://www.rae.es/}.} ({\sc drae}), the reference dictionary for the Spanish language. Their lexical categories are also checked.
\item Combination: the new resource is created from lexica that are compared and combined automatically.
\end{enumerate}

\subsubsection{Linguistic resources}
\label{sub-linguistic-resources}

We used existing Spanish linguistic resources to build {\em aLexiS}. This seemed a good start to developing a new lexicon, which required interpreting the different input data. We looked for freely available resources in terms of access, modification and distribution. We also gave importance to correctness of the entries and wide coverage. These existing linguistic resources were:

\begin{itemize}
\item {\em Lexicon of Spanish Inflected Forms}\footnote{Available at {\tt https://gforge.inria.fr/frs/?group\_id=482\&
release\_id=4290}, Nov. 2016.} ({\sc leffe}), a morphological and syntactic lexicon with wide coverage~\citep{MolineroEtAl09} based on the Alexina framework~\citep{Sagot10}. It follows the linguistic criteria applied in the equivalent lexicon for French {\sc lefff}, taking advantage of the linguistic proximity between Spanish and French as Romance languages~\citep{Sagot10}.

\item {\em OSLIN-es}\footnote{Available at {\tt http://es.oslin.org/}, Nov. 2016.}, a large-scale lexicon that
includes words attested by {\sc drae} ~\citep{Janssen05,Janssen09}.
\end{itemize}

\subsubsection{Extraction and mapping}
\label{sub-combining-extraction}
\begin{figure}
\centering
\includegraphics[scale=1.1]{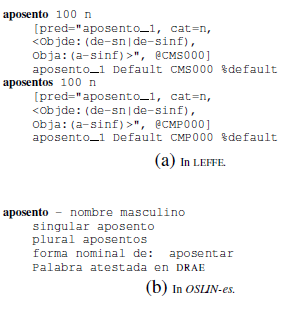}
\caption{\label{table_aposento} Example of the Spanish lemma {\em aposento} `{bedroom}' in both resources.}
\end{figure}

Extraction and mapping were performed separately; first, information was extracted from different resources and then it was mapped to a common format.
We extracted some entries from {\sc leffe} to start with. Each extracted {\sc leffe} word entry was represented in the extensional Alexina format~\citep{DanlosSagot08}. For example, Figure~\ref{table_aposento}.a represents the {\sc leffe} word entry for the Spanish lemma {\em aposento} `{bedroom}'. We chose verbs and nouns, because they play the most important role in sentences. We also extracted entries tagged as adjective, adverb, determinant, pronoun, conjunction and preposition, but discarded those tagged as interjection, numeral and proper name. 

In the example it can be observed that the lemma (after the word {\tt pred}) is a noun (represented in {\sc leffe} with {\tt C} and {\tt cat=n}) with two possible forms: masculine singular ({\tt CMS000}) for {\em aposento} `{bedroom}' and plural ({\tt CMP000}) for {\em aposentos} `{bedrooms}'.

After extraction, in order to obtain more related entries we obtained the associated lemma for each inflectional form in {\sc leffe}. Each lemma was searched automatically using the online OSLIN-es morphological database~\citep{Janssen05}. Both the forms and their morphological information and other related lemmas were retrieved. For example, Figure~\ref{table_aposento}.b shows the result of the search of lemma {\em aposento} `{bedroom}'. In this case, the retrieved word entry indicates that the lemma is a masculine noun ({\em nombre masculino}), with two forms: singular and plural. Both forms are also included in {\sc leffe}, but in this case two extra items of information were obtained: the word {\em aposento} `{bedroom}' as the nominal form of {\em aposentar} `{to lodge}', which is not a word entry in {\sc leffe}; and the searched lemma with a verified entry in the {\sc drae}. In other words, only some OSLIN-es word entries were verified. For each new word entry detected in OSLIN-es by this procedure, the process was repeated iteratively. In our example, the new search was carried out from the word {\em aposentar} `{to lodge}', yielding all the forms for the following Spanish conjugation:

\begin{itemize}
\item {\em Indicativo} `indicative': {\em presente} `present', 
{\em pret\'erito imperfecto} `past simple', {\em futuro} `future', {\em pret\'erito perfecto} `past simple', {\em condicional} `conditional';

\item {\em Subjuntivo} `subjunctive': {\em presente} `present', {\em pret\'erito perfecto} `past simple' and {\em futuro} `future';

\item {\em Imperativo} `imperative';

\item Non-finite forms: {\em infinitivo} `infinitive', {\em gerundio} `gerund' or `present participle' and {\em participio} `past participle'.
\end{itemize}

Since Spanish is a highly inflected language with about 50 conjugated forms per verb~\citep{MolineroEtAl09}, it was important to retrieve and save the complete conjugation of each verb, omitting the compound verbal forms, because these are built using the conjugation of the auxiliary verb {\em haber} `to have'. This information allowed the verbal tense to be adjusted to the semantic features of a specific context. 

The problem with the extraction process is that the word entries have different formats and tags. In order to simplify the merging step and avoid possible future inconsistencies, it was necessary to convert all entries to a common format. Algorithm~\ref{alg:stage1} describes this process.

\begin{algorithm}[H]
\small
 \caption{\label{alg:stage1}: {\bf Extraction and mapping algorithm}}
 \begin{algorithmic}[0]
 \scriptsize
  \Function{extraction\_mapping}{\{$\mbox{\sc leffe}$\}}
  \For{$e_{\mbox{\scriptsize \sc leffe}} \in \{\mbox{\sc leffe}\}$}
  \State lem$_{e_{\mbox{\scriptsize \sc leffe}}}$ = $e_{\mbox{\scriptsize \sc leffe}}$.getLemma()
  \State cat$_{e_{\mbox{\scriptsize \sc leffe}}}$ = $e_{\mbox{\scriptsize \sc leffe}}$.getCat()
  \If{lem$_{e_{\mbox{\scriptsize \sc leffe}}}$.isInOslin()}
  \State \{$e_{\mbox{\scriptsize \sc oslin}}$\} = searchInOslin(lem$_{e_{\mbox{\scriptsize \sc leffe}}}$)
  \State{\scriptsize \%Entries in Oslin with lemma morphological data.}
  \State \{$\mbox{\sc oslin}$\}.add(\{$e_{\mbox{\scriptsize \sc oslin}}$\})
  \EndIf
  \EndFor
  \State {\scriptsize \%Completion of syntactic and semantic data for new entries of {\sc oslin} not present in {\sc leffe}}.
  \For{$e_{\mbox{\scriptsize \sc oslin}} \in \{\mbox{\sc oslin}\}$}   \State lem$_{e_{\mbox{\scriptsize \sc oslin}}}$ = $e_{\mbox{\scriptsize \sc oslin}}$.getLemma()
  \State cat$_{e_{\mbox{\scriptsize \sc oslin}}}$ = $e_{\mbox{\scriptsize \sc oslin}}$.getCat()
  \EndFor  
  \EndFunction
   \end{algorithmic}
\end{algorithm}
\begin{algorithm}[H]
\small
 \caption{\label{alg:stage2}: {\bf Verification algorithm}}
 \begin{algorithmic}[0]
 \scriptsize
\vspace{-0.05cm}
  \Function{verification}{\{$\mbox{\sc set}$\}}
  \For{$e_{\mbox{\scriptsize \sc set}} \in \{\mbox{\sc set}\}$}
  \State lem$_{e_{\mbox{\scriptsize \sc set}}}$ = $e_{\mbox{\scriptsize \sc set}}$.getLemma()
  \State cat$_{e_{\mbox{\scriptsize \sc set}}}$ = $e_{\mbox{\scriptsize \sc set}}$.getCat()
  \If{!lem$_{e_{\mbox{\scriptsize \sc set}}}$.isIn{\mbox{\sc drae}}()$\ \mbox{\sc or}\ 
$!lem$_{e_{\mbox{\scriptsize \sc set}}}$.catIn{\mbox{\sc drae}  
  }(cat$_{e_{\mbox{\scriptsize \sc set}}})$\\\hspace{0.85cm}}
  \State \{$\mbox{\sc set}$\}.delete(e$_{\mbox{\scriptsize \sc set}})$
  \EndIf
  \EndFor
  \EndFunction
   \end{algorithmic}
\end{algorithm}
\begin{algorithm}[H]
\scriptsize
 \caption{\label{alg:building}: {\bf Lexicon building algorithm}}
 \begin{algorithmic}[0]
   \State \{$\mbox{\sc oslin}$\} = \{$\emptyset$\}
  \State \{$\mbox{\sc leffe}$\}= LoadLeffe()
  \State $\mbox{{\sc extraction\_mapping}(\{{\sc leffe}\})}$
  \State \{$\mbox{\sc sets}$\} = \{\{$\mbox{\sc leffe}$\},\{$\mbox{\sc oslin}$\}\}
  \For{\{set\} $\in$ \{$\mbox{\sc sets}$\}}
  \State $\mbox{{\sc verification}(\{set\})}$
  \EndFor  
  \State \{$\mbox{\sc aLexiS}$\} = \{$\mbox{\sc leffe}$\} $\cup$ \{$\mbox{\sc oslin}$\} 
 
 \end{algorithmic}
\end{algorithm}

\subsubsection{Verification}
\label{sub-combining-verification}

This step checks that the quality of word entries is satisfactory. Our solution is an automatic process that checks for the existence of each lemma and its lexical categories. In order to determine if a lemma exists and whether its lexical categories are correct, we searched for the word in {\sc drae}. We chose this for its high coverage and the fact that no training was performed, which allowed us to discard incorrect word entries and increase the precision of our lexicon. Algorithm~\ref{alg:stage2} describes the process.

\subsubsection{Merging}
\label{sub-combining-merging}
A combination step merges the collected entries using the graph unification in~\citet{necsulescu2011towards} and~\citet{BelEtAl11}. This operation is based on set unions of compatible feature values. It allows common information to be validated by adding differential data and excluding inconsistent data. It proceeds as follows:

\begin{enumerate}
\item For each common lemma, i.e. a lemma that appears in all lexica, it puts together all lexical entries with the same lemma (homography is taken into account only when lexical categories differ).
\begin{enumerate}
\item For each entry obtained in (1), a unification process is applied checking all the feature structures included in the entries.
\item Once 1.a is done, a new entry is created in {\em aLexiS}, including the set of feature structures, which contains the common information as well as any particular data in any entry from the different resources.
\end{enumerate}
\item When a lexical entry cannot be joined with any input from the other lexica, a new entry is created in {\em aLexiS} containing that unique information. The same occurs with lemmas that only belong to one lexicon.
\end{enumerate}

Note that the merging procedure avoids any possible inconsistencies thanks to the common format in the extraction and mapping step. Algorithm~\ref{alg:building} shows the sequence of steps in this section. 

\begin{figure}[!ht]
\centering
\small
\lstset{
  language=xml,
  tabsize=2,
  frame=lines,
  rulesepcolor=\color{gray},
  xleftmargin=20pt,
  framexleftmargin=15pt,
  keywordstyle=\color{blue}\bf,
  commentstyle=\color{OliveGreen},
  stringstyle=\color{black},
  numbers=left,
  numberstyle=\tiny,
  numbersep=5pt,
  breaklines=true,
  showstringspaces=false,
  basicstyle=\footnotesize,
  emph={food,name,price},emphstyle={\color{magenta}}}
\begin{lstlisting}
<Entry lemma=`aposento'>
  <feat att=`POS' val=`n'/>
  <WordForm>
    <feat att=`wForm' val=`aposento'/>
    <feat att=`gender' val=`m'/>
    <feat att=`number' val=`sg'/>
  </WordForm>
  <WordForm>
    <feat att=`wForm' val=`aposentos'/>
    <feat att=`gender' val=`m'/>
    <feat att=`number' val=`pl'/>
  </WordForm>
</Entry>

\end{lstlisting}
  \caption{\label{lexis-table} Example of Spanish lemma {\em aposento} `{bedroom}' in {\em aLexiS} entry.}
\end{figure}

Therefore, {\em aLexiS} was built by interpreting inputs extracted from previous resources, verifying them and finally merging them into a common format. Figure~\ref{lexis-table} shows the result for {\em aposento} `{bedroom}'. Since the formats of the resources were combined into this final result, the lexical information associated with the lemma and PoS tag remain the same. In the case of word forms, the information is repeated in both {\sc leffe} and OSLIN-es.

\subsubsection{Automatic lexicon extension}
\label{sec:extension-lexicon}

\begin{figure*}[!ht]
\centering
\includegraphics[scale=0.5]{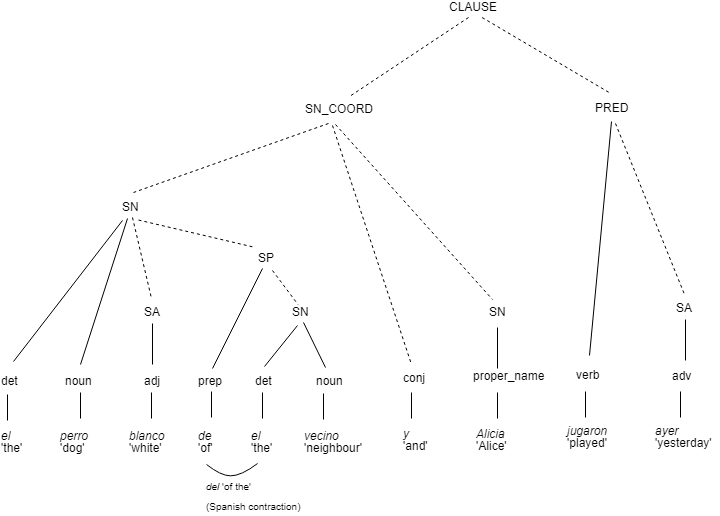}
\caption{\label{syntaxTree} Syntax tree example from the grammar.}
\end{figure*}

In order to simplify {\sc nlg} by avoiding the introduction of prepositions as input, we need to infer {\em a priori} which specific preposition follows a verb. Training was based on a dataset of Spanish novels and nearly 500 fairy tales~\citep{Andersen16,Cuentos16,Grimm16}, which had been previously {\sc pos}-tagged with Freeling Tagger\footnote{A library that provides multiple language analysis services, including probabilistic prediction of categories of unknown words~\citep{Atserias06,PadroEtAl12}.}. We developed a language model from the training process that considered bigrams and trigrams around verbs using syntactic and semantic knowledge.

Many {\sc nlg} libraries, including SimpleNLG, use collections of words without duplicates. In our case, considering the size of {\em aLexiS} and its associated linguistic data, we used an index. Moreover, by doing so our system is able to conduct the whole {\sc nlg} process more quickly.

\subsection{Syntactic structure using a grammar}
\label{grammar}

One of the challenges of computational linguistics is syntactic structuring or parsing, which consists of creating the tree parsing structure from a given sentence. We used a grammar the other way around, in order to infer the syntactic structure of the final desired sentence, for which purpose we used a {\sc 
dcg}~\citep{maggiori13}. In~\citet{gavilanes2012adquisicion}, a {\sc dcg} is defined by $ G = (N, \Sigma, P, S) $, where the formation rules are:

\begin{itemize}
\item $ \alpha A \gamma \rightarrow \alpha \omega \gamma $ \hspace{0.5cm} with $ A \in (N \cup \{S\}), \alpha, \gamma \in (N \cup \Sigma)^*, \omega \in (N \cup \Sigma)^* - \{\epsilon\} $ 

\item[] or

\item $ S \rightarrow \epsilon $
\end{itemize}

\noindent with $ |\alpha A \gamma| \preceq |\alpha \omega \gamma| $, where $ |\alpha A \gamma| $ represents the number of symbols in $ \alpha A \gamma $. The languages generated with this type of grammar are called context-sensitive languages ({\sc csl}). Given all possible tree structures within a grammar, the system selects the most appropriate structure for the input words. We chose a simple Spanish grammar allowing a wide range of basic sentences with low computational effort\footnote{{\tt http://PrologDCG-es.sourceforge.net/}.}. This has the advantage that it tries to reduce grammar rules by annotating number and person considerations. Generative grammars~\citep{ruwet1974} are context-free -- for example: {\tt sentence-->nominal syntagm, verbal syntagm} -- but, in addition to cases {\em Yo tengo fr\'io} `I am cold' and {\em nosotros tenemos fr\'io} `we are cold', they may generate the case {\em yo tenemos fr\'io} `I are cold'\footnote{This sentence is incorrect grammatically speaking due to the wrong inflection of the verb given the subject.}. In order to avoid incorrect sentences like this it is necessary to multiply the number of rules. Instead, however, the grammar of our choice is annotated with number and person features, thereby ensuring that the verb inflection is correct\footnote{This is known as an augmented grammar~\citep{grishman91}.}. The final result is: 

\begin{center}
{\tt sentence-->nominal syntagm(person, number), verbal syntagm(person, number)}.
\end{center}

\begin{figure*}[ht!]
 \centering
 \includegraphics[scale=0.6]{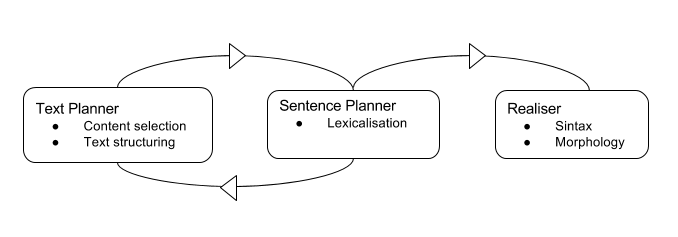}
 \caption{Our three-stage NLG architecture.}
 \label{fig:nlg}
\end{figure*}

In the next subsections the notation is as follows. Elements in upper case letters correspond to tree structures and elements in lower case letters represent word constituents, not variables. In the examples a dashed line represents non-direct substitutions as in the case of {\sc sn} within {\sc sn\_coord}. Spanish examples of the tree structures are given in italics. Figure~\ref{syntaxTree} shows a complete example of some linguistic rules extracted from the grammar.

\subsubsection{Nominal and coordinated nominal syntagm rules}

Nominal syntagms are composed of nouns, pronouns and proper names. Nouns may be preceded by a determiner or followed by an adjectival/adverbial and/or prepositional syntagm. A sentence composed of two nominal syntagms with a conjunction in between is called a coordinated syntagm.

\subsubsection{Adjectival, adverbial and prepositional syntagm rules}

In this case, a noun may be followed or not by an adjectival/adverbial syntagm. These syntagms are also composed of an adverb or an adjective that may be followed by another adjective or adverb, respectively.
As in the case of adjectival/adverbial syntagms, a prepositional syntagm may be empty or may be composed of a preposition followed by a nominal syntagm.

\subsubsection{Predicate rule}

The predicate of a sentence is composed of a verb that may be followed or not by a nominal or coordinated nominal syntagm and an adjectival/adverbial and/or prepositional syntagm. It may also be composed of a verb followed by another verb, such as in the sentence {\em Yo intento estudiar ciencias} `I try to study science'; followed by a nominal or coordinated nominal syntagm and an adjectival/adverbial and/or prepositional syntagm.

\subsubsection{Sentence rule}

A sentence is composed of a nominal coordinated syntagm followed by a predicate or a single predicate without a subject, which is very common in Spanish. The subject of the sentence may also be a nominal syntagm. Given the relations among the different syntagms, and due to computational and time limitations, we set a depth level of two iterations. Take the nominal syntagm in Figure~\ref{syntaxTree} as an example. The first nominal syntagm includes a prepositional syntagm composed of a nominal syntagm. It is easy to detect the loop condition here. In this regard, the second nominal syntagm cannot be composed of another prepositional syntagm.

\subsection{Our NLG library: surface realizer}
\label{nlg_system}

Our aim is to design an {\sc nlg} system to generate Spanish sentences from a set of input words. These input words should be meaningful, i.e. nouns, verbs and adjectives. Our system can infer determiners, prepositions and conjunctions automatically to be included in the final sentence.

Our three-stage {\sc nlg} architecture in Figure~\ref{fig:nlg} is inspired in a state-of-the-art processing pipeline, adapted for automatic Spanish {\sc nlg}. The first {\em Text Planner} stage deals with content selection and text structuring. The user selects the words to be considered when creating an {\sc nl} sentence in Spanish, in {\sc svo} (subject-verb-object) order, which is adequate in Spanish for practical {\sc nlg} applications. The system infers the text structure from the information provided by the grammar. The second {\em Sentence Planner} stage deals with lexicalization, which comprises the actions of adding words and setting phrases or word patterns. The final {\em Realizer} stage adds any extra elements needed and performs the morphology inflections to create a coherent and grammatically correct sentence in Spanish. 

The main {\sc nlg} actions are the following. In each case we indicate the stage they belong to:

\begin{itemize}

\item Detecting whether a sentence is affirmative, negative or interrogative ({\em Sentence Planner}). This decision affects the linguistic structure of the sentence. It is taken as negative if it contains the negation adverb {\em no} `not'. It is considered interrogative if it contains a question mark ({\em ?}). If the sentence contains both elements the system generates a question in negated form. In any other case the sentence is considered to be affirmative.

\item Inclusion of subject (if necessary) ({\em Sentence Planner}). Sentences with elided subjects are common in Spanish, for example {\em Voy al parque} `{I go to the park}' and {\em ?`Vais al parque?} `{Do you go to the park?}', because the inflection of the verbs allows the person and number features of the subject to be identified whether it is elided or not. Bearing in mind that we want the {\sc nlg} process to be as transparent as possible to the user, we need to minimize the number of words they must provide. When the user does not include any subject, the system takes the first singular personal pronoun {\em yo} `I' as the subject of the sentence. 

\item Syntax structure inference ({\em Text Planner}). The subject and predicate separation simplifies the search for the syntactic trees that match the input words given by the user. In this way the task is less time/resource-consuming because the trees are smaller. Thus, once the type of sentence (affirmative, negative or interrogative) is established by the {\em Sentence Planner}, the system separates the subject from the predicate according to the position of the main verb within the sentence and then looks for the best syntactic structure that fits them.

We conduct a depth-first search ({\sc dfs})~\citep{tarjan1971depth} for the best syntactic structures in our grammar given some input words. The search starts at the root and explores each branch as far as possible before backtracking. The {\sc dfs} algorithm traverses the cumulative syntactic tree in a depth-ward motion and uses a stack to remember the next vertex to start a new search when a dead end is found in any iteration.
In the example in Figure~\ref{dfs} {\sc dfs} traverses from the root retrieving paths 1-2-5, 1-2-6, 1-3, 1-4-7. It employs the following rules.

\begin{itemize}
\item Rule 1. Visit the adjacent unvisited vertex marking it as visited. Then display it. Finally, push it into the stack.

\item Rule 2. If no adjacent vertex is found, pop a vertex up from the stack. All the vertices without adjacent vertices are taken out the stack.

\item Rule 3. Repeat Rule 1 and Rule 2 until the stack is empty.

\end{itemize}

\begin{figure}[ht!]
\centering
\includegraphics[scale=0.4]{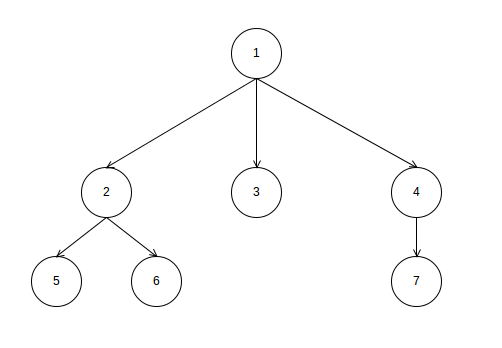}
\caption{{\sc dfs} example.}
\label{dfs}
\end{figure}

\item Addition of any extra elements needed ({\em Sentence Planner}). This is related to the previous task. Once the syntactic structure has been inferred, it may be necessary to include extra elements such as determiners, prepositions and conjunctions. These elements are included in the sentence if they correspond to feasible realizations in the grammar. This is the reason for the feedback between the first and second stages in Figure~\ref{fig:nlg}.

\item Morphological inflections ({\em Realizer}). This encompasses the inflections that are necessary to produce a sentence that is grammatically correct, in which the verb and other constituents are inflected in the way dictated by the subject. These morphological inflections not only deal with conjugation, person, number and gender features, but also with contractions and the Spanish double negation\footnote{With a second negation adverb, apart from {\em no} `not', reinforcing the negativity of the sentence. This is a notable difference with other languages like English, where a second adverb of negation makes the sentence affirmative.}. For example, the negation of the sentence {\em Yo voy siempre al teatro} `I always go to the theatre' is {\em Yo \underline{no} voy \underline{nunca} al teatro} `I never go to the theatre'. The negative meaning clearly results from the adverb {\em no} `not' and it implies changing the time adverb from {\em siempre} `always' to {\em nunca} `never'.

Since our system distinguishes between the subject and the predicate of a sentence before generating it, it can create sentences with coordinate subject applying appropriate linguistic features to each to adjust number, gender and person features. For example, considering the words {\em cuidadora, nosotros, comer, manzanas} `caregiver, we, eat, apples', the resulting sentence {\em La cuidadora y nosotros comemos manzanas} `The caregiver and we eat apples' has a compound subject.

First, gender, number and person features must be inferred for the whole sentence deriving them from the words given by the user. These features are determined by the subject, which is expected to be a nominal syntagm (coordinated or not). Continuing with the example {\em La cuidadora y nosotros comemos manzanas} `The caregiver and we eat apples', the subject is a coordinate nominal syntagm composed of two nominal syntagms. The first is composed of a feminine singular determiner {\em la} `the' and a feminine singular noun {\em cuidadora} `caregiver'. The second nominal syntagm is a pronoun {\em nosotros} `we'. Consequently, the sentence is in third person and plural. This is why the verb {\em comer} `eat' is inflected that way.

\begin{figure*}[ht!]
\centering
\hspace{0.5cm}
\includegraphics[width=0.92\textwidth]{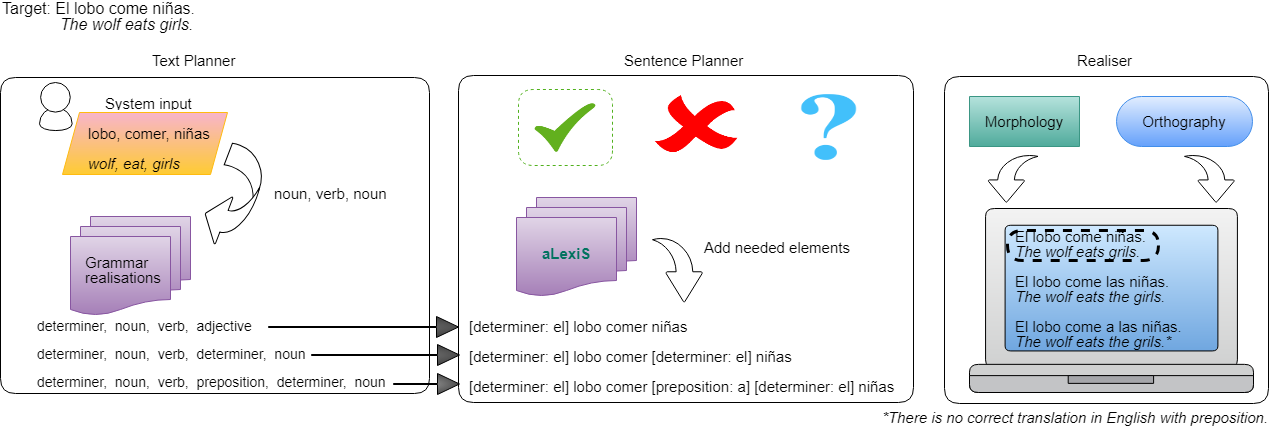}
\caption{Example of sentence generation using our three-stage {\sc nlg} architecture.}
\label{generation_diagram}
\end{figure*}

We implemented the number, gender and person linguistic rules for Spanish in {\sc drae}. In principle, the first time the user introduces the words to generate a sentence in {\sc nl}, the system takes the sentence as masculine, singular and first person, and then, using the information in {\em aLexiS}, adjusts these features applying the linguistic grammar rules. For example, if the subject is a coordinate nominal syntagm, the sentence is considered to be plural. For the gender, only if all the subject constituents are feminine the sentence is considered to be feminine. Regarding the person, it is necessary to follow the following rules in strict order to make the adjustment: 
\begin{itemize}
\item 1.
If the subject contains an element referring to the first person, the sentence is in the first person.
\item 2.
If the sentence contains an element referring to the second person that is not related to the first person, the sentence is in the second person. 
\item 3. 
If the sentence contains an element referring to the third person that is related to neither the first nor the second person, the sentence remains in the third person.
\end{itemize}

{\em aLexiS} contains the inflections of each lemma according to number and gender changes and, in the case of pronouns and verbs, also according to person features. Once these features are inferred, it is only necessary to apply them to all word inputs. Nevertheless, sometimes there is no subject included and default features should be considered (in our case, as previously mentioned, first person, masculine gender and singular number).

The verbal tense in the final sentence is present unless the user provides a time adverb. For example, if this adverb is {\em ayer} `yesterday', the sentence is in the past tense. All this linguistic knowledge is taken from {\em aLexiS}.

Our system also deals with spelling changes due to contractions, usually composed of a preposition and an article or a pronoun. For example, given the words {\em \'el, comer, con, yo} `he, eat, with, I', a contraction of the preposition {\em con} `with' with the pronoun {\em yo} `I' generates the word {\em conmigo} `with me'. So, the resulting sentence is {\em \'El come conmigo} `He eats with me'. The most common contractions in Spanish are {\em a} `to' + {\em el} `the' $\rightarrow$ {\em al} and {\em de} `of' + {\em el} `the' $\rightarrow$ {\em del}.

A fully automatic Spanish {\sc nlg} library requires default rules for atypical situations, such as input words that are not included in the lexicon. In this case they are treated as proper names. The same occurs when no related features are provided or the features cannot be inflected from the input words.

\item Orthographic rules ({\em Realizer}). We implemented the orthographic rules in {\sc drae}.

\end{itemize}

In order to generate a sentence, it is first necessary to create the syntagms that compose it and then join them paying attention to their syntactical and semantic function. For example, to generate the sentence {\em La ni\~na juega con el gato} `The girl plays with the cat', first we have to create the nominal syntagm {\em el gato} `the cat' and integrate it in a prepositional syntagm {\em con el gato} `with the cat'. It is also necessary to build the subject of the sentence {\em la ni\~na} `the girl' and the predicate with the main verb {\em jugar} `play'. Finally, the subject and the predicate have to be integrated with the prepositional complement in the final sentence. We manage all these stages automatically.

\begin{table*}[!ht]
\centering
\small
\begin{tabular}{|l|l|}
\hline
\bf Input words & \bf Best generated sentence/s\\ \hline\hline
{\em dibujar, animales} `draw, animals' & {\em Yo dibujo animales.} `I draw animals.'\\\hline
{\em Ana, ir, colegio, no} & {\em Ana no va al colegio.}\\
`Ana, go, school, not' & `Ana doesn't go to school.'\\\hline
{\em p\'ajaros, poder, volar, ?} & {\em ?`Los p\'ajaros pueden volar?}\\
`birds, can, fly, ?' & `Can birds fly?'\\\hline
{\em ni\~nas, tomar, batido, chocolate} & {\em Las ni\~nas toman el batido del chocolate.}\\& {\em Las ni\~nas toman el batido y el chocolate.}\\
`girls, have, milkshake, chocolate' & `The girls have the chocolate milkshake.'\\ & `The girls have the milkshake and the chocolate.'\\\hline
{\em profesor, escribir, letras, n\'umeros, en, pizarra} & {\em El profesor escribe las letras y los n\'umeros en la pizarra.}\\
`teacher, write, letters, numbers, on, blackboard' & `The teacher writes the letters and the numbers on the blackboard'.\\\hline
{\em abejas, volar, alrededor, de, flor, amarillo} & {\em Las abejas vuelan alrededor de la flor amarilla.}\\
`bees, fly, around, flower, yellow' & `The bees fly around the yellow flower.'\\\hline
\end{tabular}
\caption{\label{functionalitiesExample} Examples of sentences illustrating the functionalities of our {\sc nlg} library.}
\end{table*}

Figure~\ref{generation_diagram} shows another complete example of {\sc nlg} using our three-stage architecture. The target sentence is {\em El lobo come ni\~nas} `The wolf eats girls'. The user provides the system with the input words: {\em lobo, comer, ni\~nas} `wolf, eat, girls'; which are the meaningful elements in the final {\sc nl} sentence. In the first stage, the {\em Text Planner} infers three suitable linguistic realizations for the input. In the second stage, the {\em Sentence Planner} learns that the sentence is in affirmative mode and adds the extra elements according to the previously selected linguistic realizations in the grammar and the information within {\em aLexiS}. Finally, in the third stage the {\em Realizer} conducts the morphological and orthographic processes to generate one or more sentences as result.

Our system inserts conjunctions, determiners and prepositions automatically. In addition, if there is a time-related adverb like {\em ma\~nana} `tomorrow' among the word inputs, the verbal tense of the sentence is adjusted automatically (in the example, to future tense). In the special case of verbs that can be reflexive or non-reflexive, the system generates the sentence depending on the corresponding probabilities. The system gets this information from our language model in {\em aLexiS}. We developed an algorithm to build the sentence word by word based on the linguistic knowledge in the lexicon and the grammar rules that we extracted for Spanish.

\subsubsection{Functionalities}

Our library allows coherent and complete sentences in Spanish to be constructed that can be affirmative, negative or interrogative. It is possible to create simple sentences and complex sentences with compound subject or double negation. Table~\ref{functionalitiesExample} shows sentences that were created automatically in increasing order of linguistic complexity, as well as the corresponding input words.

\section{Quantitative analysis and experimental results}
\label{evaluacion}

\subsection{Lexicon}
\label{result-built}

\begin{itemize}

\item {\em GilcUB-M Dictionary}. A full form dictionary with morphosyntactic annotations in Multext encoding schema conforming with {\sc eagles} standards for morphosyntactic encoding of computational lexica. It contains 62,244 unique (lemma, PoS) pairs\footnote{According to the {\sc elra} website {\tt http://catalog.elra.info/ product\_info.php?products\_id=30}, Nov. 2016.}
(29.12\% fewer than {\em aLexiS}).

\item {\em Freeling}. A morphosyntactic lexicon used for morphological analysis and PoS disambiguation modules in the {\em FreeLing} {\sc nlp} tool~\citep{PadroEtAl10}. It uses an adapted version of the {\sc eagles} tag set and has encoded tags, such as grade for adjectives. It contains 76,318 unique (lemma, PoS) pairs 
(13.09\% fewer than {\em aLexiS}).

\item The {\em TIP Conjugator} of Spanish verbs\footnote{Available at {\tt http://tip.dis.ulpgc.es/conjugar-verbo/}, Nov. 2016.}~\citep{CarrerasEtAl10}. It provides different conjugations accepted by the Spanish academies of several geographical areas, such as in R\'io de la Plata and Canary Islands. It contains 12,862 unique (lemma, verbs) pairs (9.73\% more than {\em aLexiS}, but only has information for verbs).

\item {\em AnCora-Verb-Es}. A lexicon of Spanish verbs\footnote{Available at {\tt http://clic.ub.edu/ancora/}, Nov. 2016.}~\citep{AparicioEtAl08}. It has mappings between syntactic functions, arguments and thematic roles for each predicate. Each verbal predicate is related to one or several semantic classes differentiated according to four event classes (accomplishments, achievements, states and activities). It contains 1,965 unique (lemma, verbs) pairs (83.08\% fewer than {\em aLexiS}).

\end{itemize}

\begin{table}[ht!]
\centering
\footnotesize
\begin{tabular}{l|c|}
\cline{2-2}
& \bf Unique pairs compared to aLexiS\\\hline\hline
\multicolumn{1}{|l|}{GilcUB-M Dictionary} & 29.12\% less\\
\multicolumn{1}{|l|}{Freeling} & 13.09\% less\\
\multicolumn{1}{|l|}{TIP Conjugator} & 9.73\% more (only verbs)\\
\multicolumn{1}{|l|}{Ancora-Verb-Es} & 83.08\% less\\\hline
\end{tabular}
\caption{\label{comparison} {\em aLexiS} compared to other Spanish lexica in terms of coverage.}
\end{table}

\begin{table*}[!ht]
\centering
\small
{\begin{tabular}{{l}|{c}|{c}|{c}|{c}|{c}|{c}|{c}|{c}|}
\cline{2-9}
\multicolumn{1}{p{1cm}|}{}& \multicolumn{2}{c|}{{\bf Initial}}& \multicolumn{2}{c|}{{\bf Extracted}}& \multicolumn{4}{c|}{{\bf Extracted and Tested}}\\
\cline{2-9}
& \bf ({\em lem}, {\em tag}) & \bf ({\em form}, {\em tag}) & \bf ({\em lem}, {\em tag}) & \bf {\%lem/Ini} & \bf ({\em lem}, {\em tag}) & \bf ({\em form}, {\em tag}) & \bf {\%lem/Ini} & \bf {\%lem/Ext}\\ \hline\hline
\multicolumn{1}{|l|}{\sc leffe} & 165,000 & 680,000 & 101,920 & 61.76\% & 69,879 & 602,393 & 42.35\% & 68.56\%\\
\multicolumn{1}{|l|}{OSLIN-es} & 115,876 & 1,053,401 & 96,852 & 83.58\% & 58,743 & 778,150 & 51.63\% & 60.65\%\\\hline
\end{tabular}}
\caption{\label{table-results} {\em aLexiS} lemma and form information extraction from {\sc leffe} and {\em OSLIN-es}.}
\end{table*}

Table~\ref{comparison} summarizes the comparison between {\em aLexiS} and other Spanish lexica in terms of coverage. Table~\ref{table-results} shows the information extracted from the different resources we selected to create {\em aLexiS}.

According to~\citet{MolineroEtAl09}, {\sc leffe} contains over 165,000 unique ({\em lemma}, {\em tag}) pairs, which correspond to approximately 680,000 unique ({\em form}, {\em tag}) pairs. Taking into account that we only extracted some entries (with tags such as noun and adjective), the number of unique ({\em lemma}, {\em tag}) pairs pulled out was 101,920, of which 69,879 were tested on the {\sc drae}. This corresponds to 68.56\% of the extracted entries. The number of unique ({\em form}, {\em tag}) pairs tested on {\sc drae} was 602,393, 88.59\% of the total. OSLIN-es contains approximately 115,876 unique (lemma, PoS) pairs and 1,053,401 inflected forms (including homographic forms)\footnote{We thank Maarten Janssen for providing us with this data.}. In this case, we only extracted 96,852 unique ({\em lemma}, {\em tag}) pairs. This corresponds to 83.58\% of the lexicon. Of them, 58,743 pairs were tested on {\sc drae}, that is, 60.65\% of the extracted entries.

Table~\ref{table-results2} shows the number of lemmas and forms in {\em aLexiS} classified by lexical category. The vast majority are tagged as nouns ($\sim$ 49,200), representing over 107,000 inflected forms added to {\em aLexiS}.

As explained in~\citet{silvia2018web4all}, earlier approaches combined resources of varying formatting quality. Conversely, we chose them for their coverage and accuracy. As shown in Table~\ref{comparison}, we were able to collect more lemmas and forms by combining the selected resources than by taking information from them separately (considering only extracted and tested information).

\begin{table}[!ht]
\centering
\small
{\begin{tabular}{|{l}|{c}|{c}|}
\hline
\bf Category & \bf Lemmas & \bf Forms\\ \hline\hline
Adjective & 24,584 & 82,387\\
Adverb & 2,275 & 2,275\\
Conjunction & 37 & 37\\
Determiner & 37 & 108\\
Noun & 49,206 & 107,557\\
Preposition & 30 & 30\\
Pronoun & 32 & 76\\
Verb & 11,611 & 649,092 \\
\hline
\bf TOTAL & 87,812 & 841,562 \\\hline
\end{tabular}}
\caption{\label{table-results2} {\em aLexiS} size by lexical category.}
\end{table}

\subsection{Experimental text corpus}
\label{corpus}

We evaluated the system manually and automatically. We chose not to apply commonly used measures from the state-of-the-art, such as {\sc rouge}~\citep{lin2003automatic} and {\sc bleu}~\citep{papineni2002bleu}, among others, because they only weakly reflect human judgements of system outputs as generated by end-to-end {\sc nlg}, as supported by~\citet{novikova2017we}.

Even though it may be used for general {\sc nlg} purposes, we focused on a real application, augmentative and alternative communication within the {\em Accegal} project\footnote{Available at http://www.accegal.org/en/ .}, integrating the system with the {\em PictoDroid Lite} communicator\footnote{Available at http://www.accegal.org/en/pictodroid-lite/ .}. First we created a dataset of Spanish sentences and clauses\footnote{Available at http://www.gti.uvigo.es/index.php/en/resources/6-augmentative-and-alternative-communication-clauses-annotated-dataset-for-natural-language-generation.}. We discarded the sentences whose complexity exceeded the objectives of our communicator, like those containing subordinate clauses or explanations after a colon such as {\em All\'i estaban sus amigos: el pato, el gato y el p\'ajaro} `Their friends were there: the duck, the cat and the bird'. Of course our system could generate them, but in separate sentences (one saying who the friends were and other that they were there). We then preprocessed the result to extract the main words within the sentences (nouns, pronouns, proper names, verbs, adjectives and adverbs). Next we lemmatized all these words except for the nouns and pronouns\footnote{If we lemmatized the nouns and pronouns, the system would have no way to know that the user wants to generate a sentence containing a noun or pronoun in plural form, since the features of these words do not depend on other constituents of the sentence, as in the case of adjectives, which depend on the noun that they modify.}. The resulting dataset had 948 sentences in Spanish and the corresponding main words. 

\subsection{Evaluation procedure and results}
\label{evaluation}

Given a target sentence, we introduced its main words into the automatic {\sc nlg} system and studied the generated sentences. When the match between the target and generated sentence was total, automatic generation was considered successful. This happened with 736 sentences, 77.64\% of the total. The remaining 212 were manually inspected by 5 different annotators, all of them {\sc nlp} researchers from the {\em GTI Research Group} in {\em atlanTTic}, {\em University of Vigo}. The annotations were chosen from the options in Table~\ref{annotatedfeatures}.

\begin{table}[ht!]
\centering
\small
\begin{tabular}{|l|l|}
\hline
\bf Feature & \bf Values \\ \hline\hline
\multirow{3}{*}{Error type} & Morphological, syntactic,\\ 
& lexicon, grammar, orthographic, \\
& target, lemmatizer \\\hline
Evaluation & 0-5 \\\hline
Best generation & Optional \\\hline
Generation suggestion & Optional \\\hline
\end{tabular}                                  
\caption{\label{annotatedfeatures} Annotated features.}
\end{table}

\begin{table}[ht!]
\centering
\footnotesize
\begin{tabular}{l|c|}
\cline{2-2}
 & \bf Number of sentences \\ \hline\hline
\multicolumn{1}{|l|}{No consensus in best realization} & \multicolumn{1}{|c|}{0} \\
\multicolumn{1}{|l|}{No consensus in error with consensus} & \multirow{2}{*}{24} \\
\multicolumn{1}{|l|}{in best realization} & \\
\multicolumn{1}{|l|}{Positive annotations (total consensus)} & 188 \\\hline
\end{tabular}
\caption{\label{scenarios} Distribution of the three realization cases of our dataset.}
\end{table}

\begin{table*}[!ht]
\centering
\footnotesize
\begin{tabular}{|l|l|l|l|}
\hline
\bf ID & \multicolumn{1}{l|}{\bf Target} & \multicolumn{1}{l|}{\bf System input} & \bf Generated sentence/s \\ \hline\hline
1 & \multicolumn{1}{l|}{\em Coge el tap\'on de la botella.} & \multicolumn{1}{l|}{\em coger, tap\'on, botella} & \em Cojo el tap\'on de la botella. \\

& \multicolumn{1}{l|}{`Get the stopper of the bottle.'} & \multicolumn{1}{l|}{`get, stopper, bottle'} & {`I get the stopper of the bottle.'} \\\hline

\multirow{2}{*}{2} & \multicolumn{1}{l|}{\multirow{2}{*}{\em La ni\~na escribe en la arena.}} & \multicolumn{1}{l|}{\multirow{2}{*}{\em ni\~na, escribir, arena}} & \em La ni\~na escribe la arena. \\
& \multicolumn{1}{l|}{} & \multicolumn{1}{l|}{} & \em La ni\~na escribe con la arena. \\
 
& \multicolumn{1}{l|}{\multirow{2}{*}{`The girl writes on the sand.'}} & \multicolumn{1}{l|}{\multirow{2}{*}{`girl, write, sand'}} & {`The girl writes the sand.'} \\
& \multicolumn{1}{l|}{} & \multicolumn{1}{l|}{} & {`The girl writes with the sand.'} \\\hline
 
3 & \multicolumn{1}{l|}{\em Los cerditos ven al lobo.} & \multicolumn{1}{l|}{\em cerditos, ver, lobo} & \em Cerditos ve al lobo. \\

& \multicolumn{1}{l|}{`The piglets see the wolf.'} & \multicolumn{1}{l|}{`piglets, see, wolf'} & `Piglets sees the wolf.' \\\hline

4 & \multicolumn{1}{l|}{\em Cay\'o sal al mantel.} & \multicolumn{1}{l|}{\em caer, sal, a, mantel} & \em caer sal a mantel \\

& \multicolumn{1}{l|}{`The salt fell on the tablecloth.'} & \multicolumn{1}{l|}{`fall, salt, on, tablecloth'} & `fall salt on tablecloth' \\\hline

5 & \multicolumn{1}{l|}{\em Yo hago pis en el water.} & \multicolumn{1}{l|}{\em hacer, pis, en, water} & \em Yo hago pis en Water. \\

& \multicolumn{1}{l|}{`I pee in the toilet.'} & \multicolumn{1}{l|}{`pee, in, toilet'} & `I pee in Toilet.' \\\hline

6 & \multicolumn{1}{l|}{\em Rosa tiene ropa roja.} & \multicolumn{1}{l|}{\em rosa, tener, ropa, rojo} & \em El rosa tiene ropa roja. \\

& \multicolumn{1}{l|}{`Rosa has red cloth.'} & \multicolumn{1}{l|}{`pink, have got, cloth, red'} & `The pink has pink cloth.'\\\hline

7 & \multicolumn{1}{l|}{\em Mam\'a corta la barriga del lobo.} & \multicolumn{1}{l|}{\em mam\'a, cortar, barriga, de, lobo} & \em La mam\'a corta la barriga del lobo. \\

& \multicolumn{1}{l|}{`Mum cuts the belly of the wolf.'} & \multicolumn{1}{l|}{`mum, cut, belly, of, wolf'} & `The mum cuts the belly of the wolf.' \\\hline

8 & \multicolumn{1}{l|}{\em El pap\'a y el ni\~no pescan con la} & \multicolumn{1}{l|}{\multirow{2}{*}{\em pap\'a, ni\~no, pescar, ca\~na, en, r\'io}} & \em El pap\'a y el ni\~no pescan con la \\

& \em ca\~na en el r\'io. & & \em ca\~na en el r\'io.\\

& \multicolumn{1}{l|}{`The dad and the child fish with} & \multicolumn{1}{l|}{\multirow{2}{*}{`dad, child, fish, fishing rod, in, river'}}& `The dad and the child fish with the \\ 

& \multicolumn{1}{l|}{the fishing rod in the river.'} & & fishing rod in the river.' \\ 
 \hline\hline

9 & \multicolumn{1}{l|}{\em El pantal\'on es morado.} &{\em pantal\'on, ser, morado} & \multirow{24}{4.5cm}{\centering EXACT MATCH}\\

& \multicolumn{1}{l|}{`The trousers are purple.'} &{`trousers, be, purple'} & \multirow{9}{*}{}\\\cline{1-3}

10 & \multicolumn{1}{l|}{\em Mam\'a cepilla al perro.} & {\em mam\'a, cepillar, perro} & \\

& \multicolumn{1}{l|}{`Mum brushes the dog.'} & {`mum, brush, dog'} & \\\cline{1-3}

11 & \multicolumn{1}{l|}{\em El beb\'e empieza a caminar.} & {\em beb\'e, empezar, caminar} & \\

& \multicolumn{1}{l|}{`The baby starts to walk.'} & {`baby, start, walk'} & \\\cline{1-3}

12 & \multicolumn{1}{l|}{\em Quiero comer mel\'on y lim\'on.} & {\em querer, comer, mel\'on, lim\'on} & \\

& \multicolumn{1}{l|}{`I want to eat melon and lemon.'} & {`want, eat, melon, lemon'} & \\\cline{1-3}

13 & \multicolumn{1}{l|}{\em Mam\'a se seca el pelo con el secador.} & {\em mam\'a, se, secar, pelo, con, secador} & \\

& \multicolumn{1}{l|}{`Mum dries her hair with the dryer.'} & {`mum, dry, hair, with, dryer'} & \\\cline{1-3}

14 & \multicolumn{1}{l|}{\em Las abejas vuelan alrededor de la} & \multirow{2}{*}{\em abejas, volar, alrededor, de, flor, rosa} & \\

& \multicolumn{1}{l|}{\em flor rosa.} & & \\

& \multicolumn{1}{l|}{`The bees fly around the pink flower.'} & {`bees, fly, around, of, flower, pink'} & \\\cline{1-3}

15 & \multicolumn{1}{l|}{\em El ni\~no infla un globo gigante de} & {\em ni\~no, inflar, un, globo, gigante, de,} & \\

& \multicolumn{1}{l|}{\em color azul.} & {\em color, azul}& \\

& \multicolumn{1}{l|}{\multirow{2}{*}{`The child inflates a giant blue ballon.'}} & {`child, inflate, a, balloon, giant, of, colour,} & \\

& & {blue'} & \\\cline{1-3}

16 & \multicolumn{1}{l|}{\em El libro y el estuche est\'an dentro} & \multirow{2}{*}{\em libro, estuche, estar, dentro, de, mochila} & \\

& \multicolumn{1}{l|}{\em de la mochila.} & & \\

& \multicolumn{1}{l|}{`The book and the pencil case are} & \multirow{2}{*}{`book, pencil case, be, inside, of, schoolbag'} & \\

& \multicolumn{1}{l|}{inside the schoolbag.'} & & \\\cline{1-3}

17 & \multicolumn{1}{l|}{\em Los ni\~nos pintan con un l\'apiz azul} & {\em ni\~nos, pintar, un, l\'apiz, azul, en, papel,} & \\

& \multicolumn{1}{l|}{\em en papel blanco.} & {\em blanco}& \\

& \multicolumn{1}{l|}{`The children paint with a blue pencil} & {`children, paint, a, pencil, blue, on, paper,} & \\
& \multicolumn{1}{l|}{on the white paper.'} & {white'}& \\
\hline

\end{tabular}
\caption{\label{results} Example of sentences that were generated automatically by our system, compared with the targets.}
\end{table*}

\subsubsection{Error type}

We considered the six error types in Table~\ref{annotatedfeatures}:
\begin{itemize}
\item Morphological error (a): the gender and/or number and/or person features of one or more words of the sentence were not correctly inflected. Sentence 1 in Table~\ref{results} is an example, where the verb {\em coger} `pick up' is not correctly inflected.

\item Syntactic error (b): the sentence had missing elements such as conjunctions and prepositions (sentence 2 in Table~\ref{results}).

\item {\em aLexiS} error (c): one or more words of the sentence were not in {\em aLexiS}, so the system treated them as proper names. As a result the inflection and other actions were not correct. Note that if a word is considered a proper name, its first letter is a capital letter (sentence 3 in Table~\ref{results}). 

\item Grammar error (d): the complexity of the target sentence exceeded the capabilities of our grammar. For example, those without a {\sc svo} structure (sentence 4 in Table~\ref{results}). In this case the system simply repeated the input words.

\item Target error (e): the target was not correct (sentence 5 in Table~\ref{results}).

\item Lemmatizer error (f): the lemmatizer did not extract the input words correctly and consequently our system was unable to generate the sentence. For instance, the lemmatizer incorrectly extracted the colour pink in sentence 6 of Table~\ref{results} as the subject instead of the proper name {\em Rosa}.
\end{itemize}

\subsubsection{Rating}

The annotators rated the quality of the generation according to the following scale:

\begin{itemize}
\item 0: the sentence was not generated. The system simply repeated its main elements, as in sentence 4 in Table~\ref{results}.

\item 1: the meaning of the generated sentence was far from the target (sentence 6 in Table~\ref{results}).

\item 2: the information in the target could be understood from the generated sentence (sentence 1 in Table~\ref{results}).

\item 3: the differences between the target and the generated sentences were minimal and they did not affect the meaning (sentence 2 in Table~\ref{results}).

\item 4: the differences between the target and the generated sentences were limited to punctuation or determiners (sentence 7 in Table~\ref{results}).

\item 5: the target and generated sentences were exactly the same (sentence 8 in Table~\ref{results}). This rating was assigned automatically in the case of success.
\end{itemize}

\subsubsection{Best generation}

If the system offered several possibilities the annotator was requested to choose the best, as for sentence 2 in Table~\ref{results}, for which the best realization was the third generated sentence.

\subsubsection{Suggestion of a result}

Inspecting the errors we noticed that most of them were related to {\sc svo} ordering and, except for that, the system could have generated the target sentences. The annotators were asked to suggest generation alternatives in that case. A possible suggestion for sentence 5 in Table~\ref{results} could be {\em La sal cay\'o en el mantel} `The salt fell on the tablecloth'.

\subsubsection{Annotation results}
\begin{figure}
\centering
\includegraphics[scale=0.35]{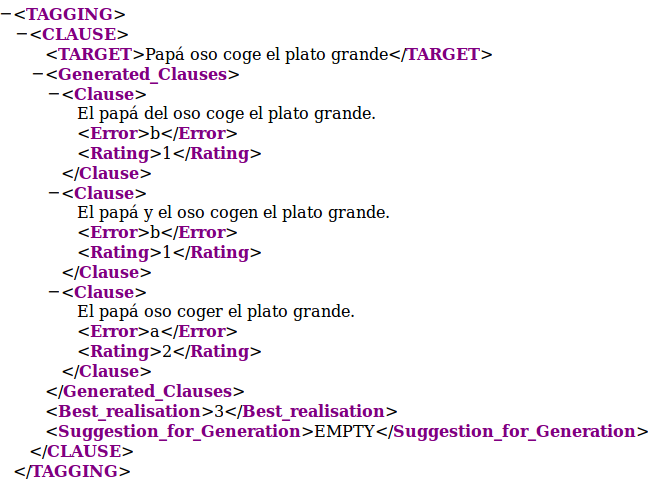}
\caption{\label{completeexample} Annotation example.}
\end{figure}

The annotation task took one month. In order to ensure the consistency of the resulting corpus, we provided some guidelines and examples to the annotators in advance. The annotation script produced an {\sc xml} file with their results. Figure~\ref{completeexample} shows an example of a completed annotated sentence.

The final results were summarized as follows. First we distinguished between the cases in which our {\sc nlg} system returned a single possibility and those with several generated sentences. In the first case we tagged the error type as that indicated by the majority of the annotators. Otherwise, we tagged the sentence with {\em no consensus in error type}. We calculated the final rating of each generated sentence as the arithmetic average of the ratings by the five annotators.
In the second case we first looked for a consensus in the {\em best\_realization} field. If there was no consensus, we tagged the sentence with {\em no consensus in best realization}. If best realizations were proposed in the second case and there was consensus among the annotators, we tagged and rated the best candidate generated by the system as in the first case. Table~\ref{scenarios} shows the distribution of the cases for our dataset.

When the annotators agreed in the second case in best realization and error, the average rating of the annotated sentences was 2. This also happened when there was no consensus in error but the annotators coincided in the best realization. The annotators made 238 different generation suggestions, of which 89 were correctly generated by our library (37.39\%).

Note that the tests covered various features of Spanish grammar such as different types of sentences (affirmative, negative, interrogative, coordinate, passive, etc.), the entire Spanish verb conjugation and constructions with different categories of words (adjectives, nouns, pronouns, etc.). Table~\ref{results} shows some examples of generated sentences. The first eight correspond to failures of our system that we have used as examples in this section. The rest are correct generations that illustrate the functionality of the system.

\subsection{Agreement measures}

Manual evaluation was assessed with two well-known agreement measures to obtain a robust estimate of the differences between the annotators: Krippendorff's {\em Alpha}-reliability and accuracy.

Krippendorff's {\em Alpha}-reliability (expression~\ref{EQalpha}) \citep{krippendorff2012content, krippendorff2011computing} is a reliability coefficient that measures agreement among observers. It analyses whether the resulting data can be trusted to represent something real. Unlike other specialized coefficients, $Alpha$ is a generalization of several known reliability indices. It allows researchers to judge a variety of data with the same reliability standard. It is valid for any number of annotators, can be applied to different variable types and metrics (e.g. nominal, ordered, interval, etc.) and can handle small or large sample sizes and incomplete/missing data. 

\begin{equation}
\centering
\small
\label{EQalpha}
Alpha = 1 - \frac{D_o}{D_e}
\end{equation}

\noindent where $D_o$ is the observed disagreement between the annotators and $D_e$ is the disagreement expected by chance rather than attributable to the properties of the coded units. When the annotators agree perfectly $Alpha = 1$, and when their agreement level seems by chance $Alpha = 0$. For the data generated by any method to be reliable, $Alpha$ should be far from this extreme, ideally $Alpha=1$. Expressions~\ref{EQdo} and~\ref{EQde} define the two disagreement measures.

\begin{equation}
\centering\small
\label{EQdo}
\displaystyle D_o = \frac{1}{n}\sum_{c}\sum_{k}o_{ck}\cdot\delta^2_{c,k}
\end{equation}

\begin{equation}
\centering\small
\label{EQde}
\displaystyle D_e = \frac{1}{n(n-1)}\sum_{c}\sum_{k}n_c\cdot n_k \cdot \delta^2_{c,k}
\end{equation}

\noindent where entities $o_{ck}$ , $n_c$, $n_k$ and $n$ refer to the frequencies of values in coincidence matrices. The first is calculated as follows:

\begin{equation}
\centering\small
\displaystyle o_{ck} = \sum_{n}\frac{\textit{Number of c - k pairs in sentence u}}{\textit{Number of annotators-1}}
\end{equation}

\begin{table}[ht]
\centering
\scriptsize
\begin{tabular}{{c}|{c}{c}{c}{c}{c}|{c}}
& \multicolumn{5}{c|}{k} & $\sum$\\
\hline
\multirow{4}{0.5cm} c & . & .& . & . & .\\
& . & .& . & . & . & . \\
& & & $ o_{ck} $ & & & $ n_{c} $ \\
& . & .& . & . & . \\
& . & .& . & . & . \\
\hline
$\sum$ & \multicolumn{5}{c|}{$ n_{k} $} & n
\end{tabular}
\caption{\label{matriz}Coincidence matrix from two different annotators into a $k \times k$ square matrix.} 
\end{table}

\noindent Expression~\ref{EQdelta} defines the difference function $\delta$.

\begin{equation}
\centering\small
\label{EQdelta}
\delta_{c,k} = \left\{ \begin{array}{lcc}
       0 & if & c=k \\
       1 & if & c \neq k \\
       \end{array}
  \right.
\end{equation}

Our evaluation scenario focused on nominal data because we measured the agreement in errors (a, b, c, d, e, f) of five observers with no missing data from our dataset. The first step was to build a reliability data matrix, a 5-observers-by-229-sentences matrix\footnote{Our system generated 229 sentences for the 212 target sentences in the corpus because there were several generated candidates for some targets.}, containing $ 5\times 229 $ values (c and k respectively).

\begin{table}[ht!]
\centering\footnotesize
\begin{tabular}{|ll|ccccccc|}
\hline
\multicolumn{2}{|l||}{\multirow{2}{*}{\bf Reliability data matrix}} & \multicolumn{7}{c|}{\bf Sentences} \\
\cline{3-9}
\multicolumn{2}{|l||}{} & \bf 1 & \bf 2 & \bf 3 & \bf ... & \bf \bf 125 & \bf ... & \bf 229 \\ \hline\hline
\multirow{5}{*}{\bf Observers} & \multicolumn{1}{|c||}{\bf Annot. 1} & b & a & d & ... & c & ... & d \\
& \multicolumn{1}{|c||}{\bf Annot. 2} & b & a & d & ... & c & ... & d \\
& \multicolumn{1}{|c||}{\bf Annot. 3} & b & a & a & ... & e & ... & d \\
& \multicolumn{1}{|c||}{\bf Annot. 4} & b & a & a & ... & f & ... & d \\
& \multicolumn{1}{|c||}{\bf Annot. 5} & b & a & a & ... & c & ... & d \\\hline
\end{tabular}
\caption{\label{reliabilityMatrix}Reliability data matrix of our annotated dataset.} 
\end{table}

The second step was to tabulate coincidence in units (Table~\ref{coincidenceMatrix}). Coincidence matrices account for all values contained in a reliability data matrix. They differ from the familiar contingency matrices, which account for units in two dimensions, not values. Our coincidence matrix tabulated all pairable errors from the five different annotators into a 6-by-6 square matrix. A coincidence matrix omits references to annotators. It is symmetric with respect to its diagonal, which contains all the perfect matches. Note that coincidences are counted twice in the coincidence matrix. Disagreements (represented by off-diagonal cells) are also counted twice, but in different cells. Table~\ref{coincidenceMatrix} shows the form of our coincidence matrix. 

\begin{table}[!ht]
\centering\small
\begin{tabular}{|c|cccc|}
\hline
\multicolumn{1}{|l|}{\bf Errors} & \bf a & \bf b & \bf ... & \bf f \\ \hline\hline
\bf a & 199.6 & 7.6 & ... & 0.8 \\
\bf b & 7.6 & 182.4 & ... & 2.8 \\
\bf ... & ... & ... & ... & ... \\
\bf f & 0.8 & 2.8 & ... & 11.2\\\hline
\end{tabular}
\caption{\label{coincidenceMatrix}Coincidence matrix of our annotated dataset.} 
\end{table}

\begin{table}[ht]
\centering\footnotesize
\begin{tabular}{|l|llll|}
\hline
\bf Agreement measure & \bf Value & & & \\ \hline\hline
\bf $Alpha$ & 0.598 & & & \\
\bf {\em Accuracy} & 0.689 & & & \\\hline
\end{tabular}
\caption{\label{alphaaccuracyresults} Overall inter-annotator agreement considering the five annotators.}
\end{table}

\begin{table*}[!ht]
\centering\small
\begin{tabular}{|ll|ccccc|}
\cline{3-7}
\multicolumn{2}{c}{\multirow{2}{*}{}} & \multicolumn{5}{|c|}{\bf Observers} \\\cline{3-7}
\multicolumn{2}{c|}{} & \multicolumn{1}{|c|}{\bf Annot. 1} & \multicolumn{1}{|c|}{\bf Annot. 2} & \multicolumn{1}{|c|}{\bf Annot. 3} & \multicolumn{1}{|c|}{\bf Annot. 4} & \multicolumn{1}{|c|}{\bf Annot. 5} \\\hline\hline
\multirow{5}{*}{\bf Observers} & \multicolumn{1}{|c||}{\bf Annot. 1} & \multicolumn{1}{|c|}{-} & \multicolumn{1}{|c|}{\textbf{0.755}} & \multicolumn{1}{|c|}{0.501} & \multicolumn{1}{|c|}{0.564} & \multicolumn{1}{|c|}{\textbf{0.646}} \\
\cline{2-7}
 & \multicolumn{1}{|c||}{\bf Annot. 2} & & \multicolumn{1}{|c|}{-} & \multicolumn{1}{|c|}{0.575} & \multicolumn{1}{|c|}{0.570} & \multicolumn{1}{|c|}{\textbf{0.602}} \\
 \cline{2-2}\cline{4-7}
 & \multicolumn{1}{|c||}{\bf Annot. 3} & & & \multicolumn{1}{|c|}{-} & \multicolumn{1}{|c|}{\textbf{0.616}} & \multicolumn{1}{|c|}{0.578} \\
 \cline{2-2}\cline{5-7}
 & \multicolumn{1}{|c||}{\bf Annot. 4} & & & & \multicolumn{1}{|c|}{-} & \multicolumn{1}{|c|}{0.561} \\
 \cline{2-2}\cline{6-7}
 & \multicolumn{1}{|c||}{\bf Annot. 5} & & & & & \multicolumn{1}{|c|}{-} \\\hline
\end{tabular}
\caption{\label{alphaMatrix}$Alpha$ reliability between pairs of annotators.} 
\end{table*}

\begin{table*}[!ht]
\centering\small
\begin{tabular}{|ll|ccccc|}
\cline{3-7}
\multicolumn{2}{c}{\multirow{2}{*}{}} & \multicolumn{5}{|c|}{\bf Observers} \\
\cline{3-7}
\multicolumn{2}{c|}{} & \multicolumn{1}{|c|}{\bf Annot. 1} & \multicolumn{1}{|c|}{\bf Annot. 2} & \multicolumn{1}{|c|}{\bf Annot. 3} & \multicolumn{1}{|c|}{\bf Annot. 4} & \multicolumn{1}{|c|}{\bf Annot. 5} \\\hline\hline
\multirow{5}{*}{\bf Observers} & \multicolumn{1}{|c||}{\bf Annot. 1} & \multicolumn{1}{|c|}{-} & \multicolumn{1}{|c|}{\textbf{0.812}} & \multicolumn{1}{|c|}{0.616} & \multicolumn{1}{|c|}{0.668} & \multicolumn{1}{|c|}{\textbf{0.734}} \\
\cline{2-2}\cline{3-7}
 & \multicolumn{1}{|c||}{\bf Annot. 2} & & \multicolumn{1}{|c|}{-} & \multicolumn{1}{|c|}{0.664} & \multicolumn{1}{|c|}{0.664} & \multicolumn{1}{|c|}{\textbf{0.690}} \\
 \cline{2-2}\cline{4-7}
 & \multicolumn{1}{|c||}{\bf Annot. 3} & & & \multicolumn{1}{|c|}{-} & \multicolumn{1}{|c|}{\textbf{0.707}} & \multicolumn{1}{|c|}{0.672} \\
 \cline{2-2}\cline{5-7}
 & \multicolumn{1}{|c||}{\bf Annot. 4} & & & & \multicolumn{1}{|c|}{-} & \multicolumn{1}{|c|}{0.664} \\
 \cline{2-2}\cline{6-7}
 & \multicolumn{1}{|c||}{\bf Annot. 5} & & & & & \multicolumn{1}{|c|}{-} \\\hline
\end{tabular}
\caption{\label{accurayMatrix}Accuracy measures between pairs of annotators.} 
\end{table*}

We next estimated the agreement between pairs of annotators with the accuracy indicator. This is defined in terms of the observed disagreement $D_o$, as shown in Equation~\ref{eqAccuracy}:
\begin{equation}
\small
\label{eqAccuracy}
\displaystyle Accuracy = 1 - D_o
\end{equation}

The accuracy is simply the average of the proportions given by the diagonal elements of the coincidence matrix. Note that it neither accounts for (dis)agreement by chance nor for the ordering of possible values. Table~\ref{alphaaccuracyresults} shows promising overall $Alpha$ and accuracy results, which are even better in Tables~\ref{alphaMatrix} and~\ref{accurayMatrix}, representing agreement by pairs of annotators (consider as a reference the inter-agreement measures in \citet{dorussen2005assessing},~\citet{poesio2005reliability} and~\citet{pestian2012sentiment}, for example).

\subsection{Comparison with the automatic SimpleNLG version}
\label{simpleNLG}

We are not aware of any other system or library that performs {\sc nlg} automatically. Therefore, we took the enhanced (automatic) Spanish version of our SimpleNLG library as a reference.

We first built a manual Spanish version of the SimpleNLG library by writing new code to satisfy the linguistic requirements of Spanish. This adaptation also uses the complete and reliable {\em aLexiS} lexicon with Spanish morphology as the basis to generate sentences. The enhanced automatic version uses Elsa, which contains not only morphological data (like {\em aLexiS}) but also syntactic and semantic information (because this version does not use a grammar). 

In the manual version, to generate a sentence it is first necessary to create the syntagms that compose it and then join them paying attention to their syntactic and semantic function. In order to generate the sentence {\em El lobo come a la abuela} `The wolf eats the grandmother', we have to create the nominal syntagm {\em la abuela} `the grandmother' and integrate it into a prepositional syntagm {\em a la abuela} `to the grandmother'. At the same time, it is also necessary to construct the subject of the sentence {\em el lobo} `the wolf' and the predicate with the main verb {\em comer} `eat'. Finally, we need to integrate the subject and the predicate with the prepositional complement in the final sentence. 

The enhanced version manages all these stages automatically. It follows a hybrid approach that combines the knowledge-base of Elsa with a language model, according to a statistical approach, to infer prepositions. Together with the lexical rules in the adapted library and those we implemented, the enhanced version can generate coherent and complete sentences. It inserts conjunctions, determiners and prepositions automatically. In addition, if the input words contain a time adverb, like {\em ma\~nana} `tomorrow', the verbal tense of the sentence is automatically adjusted. In the special case of verbs that can be used reflexively or non-reflexively, the system generates the sentence depending on the corresponding probabilities. The system gets this information from the language model we also created. We developed an algorithm to construct the sentence word by word based on the linguistic knowledge in the lexicon and the grammar rules that we implemented for Spanish. The algorithm relies on the morphological categories of the words and infers their possible syntactic function within the sentence by also using their semantic data. We refer the reader to~\citet{silvia2018web4all} for more detail.

\begin{table*}[ht!]
\centering
\footnotesize
\begin{tabular}{|l|l|l|}
\hline
\bf Target & \bf Using our {\sc nlg} system & \bf Using our version of {\em SimpleNLG}\\ \hline\hline
{\em El ni\~no pasea por la calle.} & {\em El ni\~no pasea en la calle.} & {\em El ni\~no pasea la calle.}\\
`The children walk through the street.' & `The children walk on the street.' & `The children walk the street.'\\\hline

{\em Cort\'e el filete con tijeras.} & {\em Corto el filete con tijeras.} & {\em Yo corto el filete con las tijeras.}\\

\multirow{2}{*}{`I cut the steak with scissors.'} & `I cut the steak with scissors.' & `I cut the steak with the scissors.'\\
& (in present and with elided subject) & \\\hline

{\em Mam\'a se seca el pelo con el secador.} & {\em Mam\'a se seca el pelo con el secador.} & {\em La mam\'a y se secan el pelo con el secador.}\\

`Mum dries her hair with the dryer.' & `Mum dries her hair with the dryer.' & `Mum dries her hair with the dryer.'\\\hline

{\em Me gusta la pandereta.} & {\em Me gusto la pandereta.} & {\em Me gusto la pandereta.}\\

`I like the tambourine.' & `I like the tambourine.' & `I like the tambourine.'\\\hline

{\em El lobo feroz sigui\'o a caperucita.} & {\em El lobo feroz sigue a Caperucita.} & {\em El lobo y feroz siguen el caperucita.}\\

`The big bad wolf followed Little Red & `The big bad wolf follows Little Red & `The wolf and fierce follow the Little Red\\

Riding Hood.' & Riding Hood.' & Riding Hood.'\\\hline

{\em Los cerditos corren r\'apido.} & {\em Cerditos corre r\'apido.} & {\em Los cerditos corren r\'apidos.}\\
`The piglets run quickly.' & `Piglets runs quickly.' & `The piglets run quickly.'\\\hline

{\em La bruja envenena la manzana.} & {\em La bruja envenena la manzana.} & {\em La bruja, envenenar y la manzana.}\\
`The witch poisons the apple.' & `The witch poisons the apple.' & `The witch, poison and the apple.'\\\hline

{\em El libro y el estuche est\'an dentro de} & {\em El libro y el estuche est\'an dentro} & \\
{\em la mochila.} & {\em de la mochila.} & {\em El libro est\'a dentro de la mochila.}\\
`The book and the pencil case are & `The book and the pencil case are & `The book is inside the school bag.'\\
inside the school bag.' & inside the school bag.' & \\\hline

\end{tabular}
\caption{\label{simplenlgcomparison} Comparison between our new {\sc nlg} system and the automatic Spanish version of {\em SimpleNLG} - examples.}
\end{table*}

\begin{table}[ht!]
\centering
\footnotesize
\begin{tabular}{|lc|ccc|}
\cline{3-5}
\multicolumn{1}{c}{\multirow{1}{*}{}} & \multicolumn{1}{l}{} & \multicolumn{3}{|c|}{\bf Enhanced {\em SimpleNLG}} \\
 \cline{3-5}
\multicolumn{1}{c}{\multirow{1}{*}{}} & \multicolumn{1}{l|}{} & \bf Correct & \multicolumn{1}{c|}{\bf Incorrect} & \multicolumn{1}{c|}{\bf Total} \\\hline\hline
\multirow{3}{*}{\bf Our library} & \multicolumn{1}{|c|}{\bf Correct} & 346 & \multicolumn{1}{c|}{390} & 736 \\
 & \multicolumn{1}{|c|}{\bf Incorrect} & 17 & \multicolumn{1}{c|}{195} & 212 \\ \cline{2-5} 
 & \multicolumn{1}{|c|}{\bf Total} & 363 & \multicolumn{1}{c|}{585} & 948\\\hline 
\end{tabular}
\caption{\label{comparison-nlg-simplenlg}Comparison between our new {\sc nlg} system and the automatic Spanish version of {\em SimpleNLG} - automatic generation success.} 
\end{table}

We compared the new library proposed in this paper with the automatic Spanish version of SimpleNLG that we created using the dataset in Section~\ref{corpus}. Tables~\ref{simplenlgcomparison} and~\ref{comparison-nlg-simplenlg} show the comparison. The new {\sc nlg} library outperformed the automatic version of SimpleNLG. The former generate 77.64\% of the dataset sentences, but the latter only succeeded in generating 38.29\% of them. Besides, the new library generated 390 sentences that our automatic version of SimpleNLG was unable to create. The automatic version of SimpleNLG was better for only 17 sentences.

The performance gap seems due to the difficulty to introduce new realizations in the automatic SimpleNLG version, since these must be codified within the library. However, using our new {\sc nlg} library and a grammar, a new realization is a new linguistic tree that is independent from the code of the library. This explains why, for many target sentences, the output of our automatic SimpleNLG version is considered incorrect due to the presence of an article, while the {\sc nlg} library generates several variants for a single target.

\section{Conclusions}
\label{conclusions}
Spanish {\sc nlg} is a promising field where automation and flexibility deserve attention. We have described an alternative to the common {\sc nlg} practice of using templates in some steps of the generation process~\citep{ramos2016role}. Relying on the {\em aLexiS} lexicon and our grammar, our system performs fully automatic {\sc nlg} from a minimum set of words. Additional lexical resources can be easily integrated thanks to the automatic building process. The architecture allows domain-dependent components to be separated from domain-independent components, which can be reused or substituted. In this regard, the {\em aLexiS} lexicon and the grammar are interesting results in themselves that could be useful for many other {\sc nlg} systems or applications. Besides, the system could also be used to check for grammatical errors.

To the best of our knowledge this is the first attempt to create a fully automatic system for Spanish {\sc nlg} following a hybrid approach, that is, combining a knowledge-based approach (linguistic data, i.e. vocabulary and grammar realizations) and a machine learning approach (language model). Our system is able to generate complete, coherent and correctly spelled sentences from the main word sets supplied by the user. Noteworthy is the easy extension of our system to other languages due to its modular design and implementation, focused on integrability, portability and efficiency for feasible integration in a wide range of digital devices.

It may be argued that modularity is compromised by the vocabulary and the grammar required to perform the {\sc nlg} task in an automatic way. Indeed, the system relies on this data to take most of the surface realization decisions. For this reason, we have provided a detailed explanation of the lexicon creation procedure and the grammar used. We are aware of the fact that our library expects the user to introduce the input words respecting an {\sc svo} order (broadly tolerated in the {\sc nlg} field). In this regard, a future line of research is predicting the best grammar realization for a set of non-ordered input words to further improve the performance of our system.

Automatic and manual evaluation resulted in a positive experimental outcome. We applied state-of-the-art metrics as well as human supervision to assess system performance. The system was able to automatically and correctly generate 77.64\% of the sentences. Due to the novelty of our automatic approach and the lack of datasets for similar tasks, we had to create an automatic version of other library from the literature as a competitor (see Section \ref{simpleNLG}). 
%4.5.). 
According to this comparison, we obtained a significant 35.35\% accuracy improvement. The evaluations of other state-of-the-art {\sc nlg} solutions seek an exact match between the target and generated texts. Conversely, we performed an insightful analysis of similarities and main differences at a semantic level. In the {\sc aac} use case, our system may help people with communication impairments to express themselves in an easier, faster and more natural way.

As future research lines, first we will adapt our system to use by other languages. Second, we will study to what extent our system may help people with communication disabilities. Last but not least, as previously said, we seek to predict the best grammar realization for a set of non-ordered input words to further improve performance.

\section*{Acknowledgements}
This work was partially supported by Mineco grant TEC2016-76465-C2-2-R and Xunta de Galicia grants GRC 2014/046 and ED341D R2016/012.

\bibliography{mybibfile}

\end{document}